\documentclass{article}

\usepackage[final]{neurips_2023}
\usepackage{natbib}
\usepackage{multicol}
\usepackage[edges]{forest}
\usetikzlibrary{shadows}
\usepackage[symbol]{footmisc}

\usepackage{comment}
\usepackage[utf8]{inputenc} 
\usepackage[T1]{fontenc}    
\usepackage{hyperref}       
\usepackage{url}            
\usepackage{booktabs}       
\usepackage{amsfonts}       
\usepackage{nicefrac}       
\usepackage{microtype}      
\usepackage{xcolor}
\usepackage{tikz}
\usepackage{relsize}
\newcommand{\smallts}[1]{\textsuperscript{\scalebox{1}{#1}}}

\newcommand\lword[1]{\leavevmode\nobreak\hskip0pt plus\linewidth\penalty50\hskip0pt plus-\linewidth\nobreak #1}
\newcommand\inlinecode[2][]{\lword{\tikz[overlay]\node[fill=blue!20,inner sep=1pt, anchor=text, rectangle, rounded corners=1mm,#1] {\ttfamily #2};\phantom{\ttfamily #2}}}

\title{Leveraging AI for 
Natural Disaster Management : Takeaways From The Moroccan Earthquake}

\author{\textbf{Morocco Solidarity Hackathon} \thanks{Web site: \href{https://morocco-solidarity-hackathon.io}{Morocco Solidarity Hackathon}} \\ \textbf{Organizers, Speakers, Mentors and Participant teams}}

\begin{document}

\maketitle

\begin{abstract}

The devastating 6.8-magnitude earthquake in Al Haouz, Morocco in 2023 prompted critical reflections on global disaster management strategies, resulting in a post-disaster hackathon, using artificial intelligence (AI) to improve disaster preparedness, response, and recovery. This paper provides (i) a comprehensive literature review, (ii) an overview of winning projects, (iii) key insights and challenges, namely real-time open-source data, data scarcity, and interdisciplinary collaboration barriers, and (iv) a community-call for further action.
\end{abstract}
  
\vspace{-2mm}
\section{Introduction}
\vspace{-3mm}
Natural disasters, including earthquakes, wildfires, and floods first impact the most vulnerable populations. On September 8th, 2023, a 6.8-magnitude seism hit Al Haouz, Morocco, causing 2,946 fatalities, 5,674 injuries, 50,000 damaged homes, in particular in the most vulnerable regions of the Atlas Mountains, and damaged heritage sites in the region~\citep{imc,disPhil,britannica}. A group of students and researchers felt compelled to act, and they organized a hackathon one week later. They invited the broader community to deliberate on the use of Artificial Intelligence (AI) to monitor and mitigate such natural disasters. A unifying theme emerged: the importance of obtaining extensive, real-time, and open-source data to amplify their societal impact.

\vspace{-2mm}
\section{Literature Review}
\label{gen_inst}
\vspace{-3mm}

In what follows, we delve into AI's role, particularly machine learning (ML), in natural disaster management, drawing from numerous studies~\citep{ MR2, MR3, MR4, Dineva2023, chamola2020disaster}, insights from hackathon speakers, and suggesting potential directions for future mid- to long-term research. This review does not encompass all aspects or areas addressed by winning hackathon teams, which focused on more pressing and immediate needs in response to the community's call. Instead, it serves as an initial snapshot of existing work, with a comprehensive and systematic review to be conducted in the future.
AI algorithms improve forecasting, early warnings, and disaster response, assisting decision-making and resource allocation through comprehensive data analysis for natural disaster risk assessment, and climate change adaptation and mitigation. ML techniques are used to analyze large datasets and make accurate predictions
for effective disaster management. However, challenges persist, including diverse AI methods and hazards, intricate data collection and handling, and decision-makers' willingness to act.
AI approaches for natural disasters aligns with several United Nations Sustainable Development Goals (SDGs), including SDG 11 (Sustainable Cities and Communities), SDG 13 (Climate Action), and SDG 15 (Life on Land)~\citep{Dineva2023,Tan2020,MR3}. These approaches enhance prediction and resource distribution reducing disaster's impacts~\citep{Dineva2023,chamola2020disaster,MR4}. Green AI contributes to intelligent, sustainable urban ecosystems, i.e. SDG 11~\citep{Tan2020,verdecchia2023systematic}. AI's climate science explorations, driven by ML, bolster adaptation and mitigation strategies related to SDG 13~\citep{Dineva2023,Tan2020}. AI's impact is also significant in health and well-being (SDG 3), minimizing disaster-related health consequences~\citep{chamola2020disaster}. However, AI's effectiveness varies among disaster types~\citep{Dineva2023}, deployment requires strict governance and regulation to manage risks~\citep{Tan2020}, and joint multi-stakeholder efforts are required for responsible AI use.

\textbf{Data collection and preparation.}
Data collection and processing are crucial for AI-based natural disaster management relying on ML techniques. High-quality datasets encompass diverse data sources, including satellite imagery, remote sensing~\citep{ivic2019artificial}, seismic activity, meteorological~\citep{Velev2023}, and geospatial data~\citep{kia2012artificial} for natural disaster risk assessment. Data fusion techniques integrate multiple sources for a comprehensive view of disaster-related factors~\citep{arfanuzzaman2021a}. Preprocessing and feature engineering are vital for cleaning, transforming, and extracting relevant features~\citep{sun2020a}, ensuring data quality for ML model effectiveness~\citep{kuglistsch2022facilitating}. Data quality and quantity pose numerous challenges, including issues with satellite imagery resolution, meteorological data gaps during rare extreme weather events, and absence of historical data~\citep{ivic2019artificial}. Addressing said challenges calls for data calibration, correction, resolution improvements, sufficiency, and representation~\citep{kuglistsch2022facilitating}. Integrating satellite and drone imagery, mobility, and social media data can improve hazard models, enabling accurate predictions for disasters like floods, fires, and earthquakes~\citep{gevaert2021fairness, gfdrr2018machine, burke2019thermal, syifa2019artificial, duarte2018satellite}.

\textbf{Natural disaster preparedness.} Natural disasters cause significant damage; accurate forecasting is crucial for establishing preparedness strategies and efficient disaster response plans. ML improves earthquake prediction and post-disaster assessment. Random forests (RFs) outperform physics-based modeling ~\citep{Zhu2023}, while geophysical indicators support seismic activity predictions using decision trees and SVM~\citep{Chelidze2022}. Multi-layer perceptrons (MLP) predict earthquakes based on geophysical laws~\citep{Reyes2013}, and recurrent neural networks (RNN) refine predictions with GPS data~\citep{Narayan2018}. Adaptive neuro-fuzzy inference systems (ANFIS) analyze seismic patterns, providing competitive results~\citep{AlBanna2020,Rana2015}. A hybrid one-week-ahead prediction model achieves 70\% accuracy~\citep{Saad2018}. In~\citep{Mignan2020}, a meta-analysis emphasizes the need for transparent ML models. Integrating varied datasets, rigorous validation, and interdisciplinary collaboration are pivotal for grounding ML's potential in earthquake prediction.

Climate change necessitates improved flood forecasting methods, considering rainfall-runoff, flash flood, streamflow, storm surge, precipitation, and daily outflow~\citep{mosavi2018flood}. Traditional ML algorithms, such as RFs and SVM, are used for rainfall prediction and flash flood mitigation~\citep{yu2017comparison}. ANFIS is prevalent in flood risk management, estimating peak flow and water levels~\citep{jimeno2017estimation, chang2006adaptive}. MLP and RNN models are widely applied in flood risk management and rainfall-runoff forecasting using rainfall, runoff, and evapotranspiration data~\citep{rezaeian2010daily, xiang2020rainfall}. Moreover, deep learning models trained on radar-data, with physics assimilation, successfully forecasted rainfall $12$--$24$ hours ahead~\citep{sonderby2020metnet, espeholt2022deep, andrychowicz2023deep}.

In wildfire management, fuzzy clustering and ANFIS forecast wildfires using meteorological data~\citep{jayakumar2020wildfire}, while logistic regression, RFs, and convolutional neural networks (CNN) predict wildfire spread with weather and historical fire records~\citep{huot2022next, radke2019firecast}. A comprehensive study used RFs, SVM, and MLP for detecting high-probability large wildfire events~\citep{perez2021machine}. For hurricane management, a multi-modal approach using XGBoost and encoder-decoder architecture was developed for 24-hour intensity and track forecasting~\citep{boussioux2022hurricane}. In~\citep{gao2023hurricast}, a Hybrid methodology combines k-means and ARIMA models to capture hurricane trends, and an autoencoder architecture simulates hurricane behavior. Recently, purely data-driven weather prediction models~\citep{pathak2022fourcastnet, lam2022graphcast, bi2023accurate}, have demonstrated remarkable skill in multi-day comparisons to state-of-the-art numerical weather prediction and hurricane trajectory forecasts. See~\citep{atmos11070676} for a review on ML models for cyclone forecasting.

\textbf{Disaster relief logistics and response} In the aftermath of disasters, crucial logistics decisions are needed. Mathematical optimization is a pivotal tool for guiding these decisions. We outline the most important types of relief decisions and refer readers to further studies~\citep{klibi2018prepositioning, gholami2019dynamic, bertsimas2021predictions, banomyong2019systematic, gupta2019big, kundu2022emergency}.

{\it Evacuation} is a vital strategy to protect people from the impacts of disasters.~\citep{BAYRAM201663} surveys optimization and simulation methods for fast and smooth evacuations. {\it Timely delivery of relief supplies} like food, water, medicine, and tents is essential to assist the victims. Numerous papers propose optimization methods for operational decisions~\citep{BENTAL20111177,Ihsan} and strategic positioning of humanitarian logistics centers~\citep{STIENEN2021102494}. In particular, optimizing the location of supply facilities to ensure robustness and to hedge against all uncertainties is of paramount importance \citep{balcik2008facility,doyen2012two}.
Various {\it medical aid optimization} approaches have been suggested. For instance,~\citep{JIA2007257} propose a temporary medical aid facility location model. See~\citep{BOONMEE2017485} for more examples. Rapid  transportation of casualties to emergency facilities is crucial, and many optimization addressing this need  have been proposed in the literature~\citep{FARAHANI2020787}.

While numerous optimization methods for disaster relief exist in literature, only a few have been practically implemented due to data uncertainty and scarcity~\citep{kunz2017relevance}. Acquiring accurate, real-time information from disaster areas is challenging. To address these uncertainties, various stochastic and robust data-driven decision models have been proposed (e.g.,~\citep{GHASEMI2020100745,Ihsan}). We advocate a better integration of AI and optimization: the data needed by relief optimization models can be obtained by AI in a much faster and more efficient way~\citep{boccardo2015uav}. Satellite or drone images, combined with DL-based models (e.g., CNNs, transformers), provide valuable information on \textbf{accessible roads}~\citep{zhao2022road,wu2021building} for route planning and resource allocation~\citep{daud2022applications,european_commission_2023}, flood inundation~\citep{munawar2021application}, and \textbf{building and health care infrastructure damage}~\citep{build1,build2,cheng2021deep}, guiding relief optimization models such as resource allocation and relief planning~\citep{munawar2021integrated}.
Social media data, combined with AI-driven solutions such as ML and natural language processing, can also enhance input data for disaster relief optimization models~\citep{xbd}.

\textbf{Communication and sensing technologies} In the absence or destruction of infrastructure, communication and multimodal sensing technologies are instrumental in enabling a swift response in post-earthquake management, notably for remote areas and villages. Specifically, aerial platforms can collect sensitive data on the fly and prioritize the deployment of emergency response units. Drones can be remotely controlled over ultra-reliable 5G wireless link, while taking into account harsh weather conditions, intermittent loss of connectivity or loss of line of sight visual from remote operators. In the fully autonomous setting when remote control operations are not possible, self-organizing drone swarming solutions using various sensors (LiDAR; cameras, thermal sensors, etc.) can help sense and fuse information from distributed geographical areas. This challenging problem mandates effective ways of sampling important sensory information, optimal path planning and communication over the air~\citep{9169921}. In these settings, inter-UAV communication can be enabled via various interfaces such as WiFi, free-space-optics and a network of orbiting satellites, e.g., the recent Starlink satellite used in the aftermath of the Moroccan earthquake to re-establish connectivity. For high situational awareness, immersive technologies such as VR/AR over low-latency and highly reliable 5G multi connectivity can boost network coverage and facilitate the communication and sharing of high-definition sensing information to medics or other emergency teams~\citep{9476381}. Among the plethora of sensing modalities, wireless (RF) sensors have several advantages as they can see through walls~\citep{Fadel} for rubbles by detecting reflected RF waves. Likewise, sensing and localization can be enabled by transmitting low-powered microwave/RF signals through rubble to look for reflection changes as deployed by NASA, using its FINDER program, during the Nepalese 7.8  magnitude earthquake~\citep{DHSNASA, DHS}.

\vspace{-2mm}
\section{Moroccan Solidarity Hackathon: General Vision, Timeline \& Results}
\label{headings}
\vspace{-3mm}
The Moroccan Solidarity Hackathon was initiated in response to the seismic event in Morocco on September 8th, 2023, that revealed a recurring pattern characterized by significant loss of life. This pattern predominantly stemmed from inadequate pre-disaster preparedness and the complexities surrounding effective relief and rescue operations. While the hackathon's inception was motivated by the necessity to provide tangible contributions to earthquake risk mitigation,
 its scope rapidly expanded to encompass the broader domain of natural disaster management.

The fundamental vision of the hackathon was to establish a collaborative platform for individuals driven by their commitment to mitigating the risks associated with natural disasters. To fulfill this vision, the hackathon focused on three distinct tracks: \textbf{ (1) Natural Disaster Preparedness.} Strategies and solutions to enhance preparedness measures for future natural disasters. This encompassed the improvement of forecasting capabilities for various natural events, such as hurricanes, floods, volcanic eruptions, and earthquakes. Additionally, it involved strategically placing critical infrastructure, such as hospitals and police stations, to optimize their resilience in the face of such disasters. \textbf{(2) Relief Rescue Efforts.} Optimization of relief and rescue operations during and after natural disasters, emphasizing overcoming logistical challenges. \textbf{(3) Data Curation.} The curation and utilization of crucial datasets are imperative; these datasets include for example satellite and drone imagery used for damage assessment and individually collected data employed to identify distressed communities.Teams recognized that the raw data alone was insufficient, so they also labeled data (manually or using ML), including road detection, to facilitate effective on-ground rescue efforts.

Hackathons, though valuable for generating innovative ideas quickly, often have limited impact due to their short duration. To overcome this limitation, our hackathon aimed to extend its impact beyond the event itself. It sought to engage winning teams and interested participants, nurturing the development of their ideas for real-world implementation. During the event, each team created a 3-minute presentation and a one-page report. Mentors provided guidance, and a post-hackathon jury selected winners based on a set of predefined criteria (detailed in Appendix \ref{judges_criteria}).

\vspace{-2mm}
\section{A Summary of Spotlight Projects \& Directions}
\label{others}
\vspace{-3mm}
We identify promising contributions for practical implementation, assessed by the jury based on criteria inspired by the UN's solution quality evaluation metrics (detailed in Appendix \ref{judges_criteria}).

\vspace{-1mm}
\subsection{DeepAster}
\vspace{-2mm}

DeepAster aims to leverage satellite imaging data to provide a real-time assessment of the impact of natural disasters, enabling precise resource allocation based on immediate needs. 
\newline
\textbf{Contribution.} The solution offers an interactive map showing emergency levels, including building detection, emergency degree calculation, and estimating affected population. It was fine-tuned to recognize North African-style roofs, showcasing its feasibility and potential impact.
\newline
\textbf{Challenges.} Limited access to real-time satellite images and initial challenges with model generalization due to the architectural peculiarity of Moroccan buildings. To overcome the latter, alternative data sources were used, for instance, Maxar's Morocco Earthquake Open Data Program~\citep{maxarcom2023morocco}, along with a manually labeled dataset specific to Moroccan buildings. 
\newline
\textbf{Key Takeaways and Future Work}  The team envisions refining the model's adaptability to different building structures and leveraging meta-learning to address data scarcity.

\vspace{-1mm}
\subsection{SOS Drone}
\vspace{-2mm}


The SOS Drone project explores the use of \textit{Unmanned Aerial Vehicles} (UAVs) for post-disaster response and impact on infrastructure assessment. 
\newline
\textbf{Contribution.} The project had three key impacts: analyzing the effects on strategic routes, locating and estimating survivor numbers, and evaluating the impact on buildings. A pre-trained YOLOv6 \citep{li2022yolov6} variant \citep{huggingfaceco2023humandetection} was fine-tuned for enhanced human detection in disaster scenarios, while a model inspired by MSNet architecture \citep{zhu2021msnet} augmented building damage assessment capabilities. The route analysis combined  image classification and object detection, with the integration of RDD2022 data \citep{arya2022rdd2022} anticipated to further refine road damage assessments due to earthquakes.
\newline
\textbf{Challenges.} Limited training data and regulatory drone constraints in Morocco meant that substantial computational resources were required for real-time algorithm development.
\newline
\textbf{Key Takeaways and Future Work}
The ultimate goal of the team is to ensure that the systems remain at the forefront of disaster management strategies, safeguarding lives and minimizing the disasters' impact on communities, by refining the AI algorithms and broadening the project's scope to cover various disaster scenarios and geographies.

\vspace{-1mm}
\subsection{Team-of-5}
\vspace{-2mm}

Team-of-5's project highlights the vital role of real-time mapping in disaster management. They identified a crucial need for efficient route planning for emergency services in disaster-affected areas, such as Morocco's Atlas Mountains, which face challenges due to outdated mapping systems, compromised road infrastructure, and uncoordinated volunteer efforts.
\newline
\textbf{Contribution} The project proposed a scalable solution combining high-resolution satellite imagery and crowdsourced data, focusing on identifying road damage or blockage. They explored data from many sources including Maxar Open Data Program \citep{maxarcom2023morocco}, DeepGlobe \citep{deepglobeorg2023deepglobe}, UNOSAT United Nations Satellite Centre \citep{unosatorg2023unosat}, and others, while also planning to incorporate information from ``Aji Nt3awnou'' Rescue Map (see section \ref{datacuration}) and Waze. Implementation involved state-of-the-art deep learning architectures like YOLOv8~\citep{github2023ultralyticsultralytics} and aimed to develop a WhatsApp-based service to ensure accessibility and wider adoption, even under varying connectivities.
\newline
\textbf{Challenges}
Throughout the project, the team navigated through the heterogeneous nature of geospatial data and the resource-consuming aspects of model training. These challenges highlighted the importance of cross-functional collaboration across different time zones and backgrounds, effective communication, goal setting, and applying technical skills in a new domain under time constraints.
\newline
\textbf{Key Takeaways and Future Work}
The use of diverse data sources, advanced architectures, and the significant role of centralized entities were key insights. Looking ahead, addressing data heterogeneity, optimizing computational resources, and enhancing collaboration will be focus areas. The development of a user-centric WhatsApp-based service and the application of lessons learned in collaboration and communication will guide future work in real-world emergency scenarios.

\vspace{-1mm}
\subsection{Grooming in Darija}
\vspace{-2mm}


The Grooming in Darija project unveiled a critical concern regarding the increase in sexual predation on social media following natural disasters, particularly observed after the Moroccan earthquake \citep{lane-2023-naturaldisasters}. The team worked diligently to detect predatory content in Darija, a low-resource Moroccan Arabic dialect, emphasizing the urgency of enhanced data collection mechanisms and inclusive practices. 
\newline
\textbf{Contribution} The team effectively employed pre-existing NLP models, fine-tuned on collected samples, yielding promising results; notably, DarijaBERT \citep{darijabert} achieved an accuracy of 75\% and an F1-score of 0.73. Their efforts highlighted the urgent need for enhanced data collection mechanisms and inclusive practices, especially for less prevalent languages and sensitive topics.
\newline
\textbf{Challenges} Addressing data scarcity and privacy in sensitive areas like abuse posed the primary challenge. The key takeaway underscored the grave reality of handling grooming and abuse data and the imperative need for innovative techniques that can generalize effectively with limited examples.
\newline
\textbf{Key Takeaways and Future Work}
The project stressed the importance of AI reducing its reliance on extensive data. It advocated for methods that can perform well with limited data, fostering inclusivity and tackling issues tied to various languages, cultures, and sensitive topics.

\vspace{-1mm}
\subsection{Data Curation}
\label{datacuration}
\vspace{-2mm}

Following the earthquake in Morocco, the team worked on enhancing the "Aji Nt3awnou" \footnote{\url{https://huggingface.co/spaces/nt3awnou/Nt3awnou-rescue-map}} (``Let's Help One Another'') 
 platform, a real-time interactive map fed by citizens' and NGOs' inputs. 
\newline
\textbf{Contribution} Aji Nt3awnou aims to efficiently allocate resources and support by accurately mapping and prioritizing the evolving needs of victims. This project focused on improving map UI/UX, matching villages to coordinates, developing a ranking algorithm, and dealing with multiple dialects.  
\newline
\textbf{Challenges.}
The team overcame numerous obstacles, including linguistic diversity, lack of datasets for low-resource languages, computational constraints, and challenges in scaling, underscoring the essential role of meticulous data curation and cleaning.
\newline
\textbf{Key Takeaways and Future Work}
Moving forward, the team plans to generalize their approach and create tutorials and guides, ensuring the adaptability and scalability of their solutions for future projects and diverse disaster scenarios.

\vspace{-1mm}
\subsection{ML4Quake: Early Prediction \& Warning}
\vspace{-2mm}
ML4Quake leverages INSTANCE (Italian Seismic Dataset \citep{michelini2021instance})  to improve earthquake early-warning alerts, aiming to predict quakes 10 seconds before they happen. Current alert systems trigger alerts 3 to 5 seconds post-quake\footnote{\url{https://scienceexchange.caltech.edu/topics/earthquakes/earthquake-early-warning-systems}}, underscoring the urgent need for life-saving technology advancements. 
\newline
\textbf{Contributions}
Utilizing 3-channel waveform recordings from the dataset, which included both earthquake and noise, models were trained to focus on the first 10 seconds of each 120-second recording. The application of RFs and Neural Network models yielded promising results, with RF achieving 87\% test accuracy and an F1-score of 84\%.
\newline
\textbf{Challenges}
Limited availability of comprehensive seismic waveform datasets and constraints in utilizing 120-second waveforms. Centralized entities are pivotal, providing access to extensive datasets and fostering collaborations among seismologists, emergency management organizations, and computer scientists, which is essential for building a comprehensive early warning system.
\newline
\textbf{Key Takeaways and Future Work}  The collaboration of computer scientists and seismologists is crucial to gather comprehensive seismic datasets, enabling robust early warning systems, which are crucial to save lives and improve emergency response coordination.

\vspace{-2mm}
\section{Discussion \& Takeaways}
\vspace{-2mm}
During the hackathon, speakers, mentors, and participants shared key insights, summarized in two-fold.

\textbf{(1) No Silver Bullet -- Teamwork in Disaster Management:} The hackathon experience emphasized the importance of a collaborative future, where AI/ML is employed in conjunction with expertise from climatology, engineering, data science, operations research and operations management. Partnerships among governments, NGOs, tech companies, and local communities, prioritizing robust, reasoning, and responsible systems, for social good, are essential. As Prof. Bengio highlights, ``The AI people can't solve these problems by themselves; it's always a collaboration with many different expertise, $\cdot$, many different points of views.''
\textbf{(2) Real-time data-driven disaster management:} The importance of collecting and analyzing data in real-time cannot be overstated, as the most impact occurs in first $72$ hours post-disaster. Developing resilient data acquisition systems, capable of withstanding adverse conditions, is essential. AI-assisted data labeling can further enhance risk mitigation strategies by transforming raw, unannotated datasets into valuable insights. Prof. Murphy's cautionary advice, "you've got to really think about the privacy and unanticipated ethical consequences," highlights responsible ethical AI, while addressing data fragmentation and scarcity of useful datasets.

\begin{ack}
This event was made possible thanks to the great contributions of: Ilias Benjelloun, Mohamed Amine Bennouna, Rachade Hmamouchi, Denis Luchyshyn, Yasser Rahhali, Rim Assouel, Abdellatif Benjelloun Touimi, Kristi Rhoades, Yassir El Mesbahi and Christophe Gallant. A big thanks also to the hackathon sponsors: MILA, UQAR, AWS, Montréal NewTech, Cooperathon, Videns Analytics and R2I.
\end{ack}

\newpage
\bibliographystyle{apalike}
\bibliography{Bibliography.bib}

\begin{thebibliography}{}

\bibitem[Adib and Katabi, 2013]{Fadel}
Adib, F. and Katabi, D. (2013).
\newblock See through walls with wifi!
\newblock In {\em Proceedings of the ACM SIGCOMM 2013 conference on SIGCOMM}, pages 75--86.

\bibitem[Al~Banna et~al., 2020]{AlBanna2020}
Al~Banna, M.~H., Taher, K.~A., Kaiser, M.~S., Mahmud, M., Rahman, M.~S., Hosen, A.~S., and Cho, G.~H. (2020).
\newblock Application of artificial intelligence in predicting earthquakes: state-of-the-art and future challenges.
\newblock {\em IEEE Access}, 8:192880--192923.

\bibitem[Andrychowicz et~al., 2023]{andrychowicz2023deep}
Andrychowicz, M., Espeholt, L., Li, D., Merchant, S., Merose, A., Zyda, F., Agrawal, S., and Kalchbrenner, N. (2023).
\newblock Deep learning for day forecasts from sparse observations.
\newblock {\em arXiv preprint arXiv:2306.06079}.

\bibitem[Arfanuzzaman, 2021]{arfanuzzaman2021a}
Arfanuzzaman, M. (2021).
\newblock {Harnessing artificial intelligence and big data for SDGs and prosperous urban future in South Asia}.
\newblock {\em Environmental and Sustainability Indicators}, 11:100127.

\bibitem[Arya et~al., 2022]{arya2022rdd2022}
Arya, D., Maeda, H., Ghosh, S.~K., Toshniwal, D., and Sekimoto, Y. (2022).
\newblock Rdd2022: A multi-national image dataset for automatic road damage detection.
\newblock {\em arXiv preprint arXiv: 2209.08538}.

\bibitem[Avishan et~al., 2023]{Ihsan}
Avishan, F., Elyasi, M., Yan\i{}ko\u{g}lu, I., Ekici, A., and \"{O}zener, O.~O. (2023).
\newblock Humanitarian relief distribution problem: An adjustable robust optimization approach.
\newblock {\em Transportation Science}, 57(4):1096--1114.

\bibitem[Balcik and Beamon, 2008]{balcik2008facility}
Balcik, B. and Beamon, B.~M. (2008).
\newblock Facility location in humanitarian relief.
\newblock {\em International Journal of logistics}, 11(2):101--121.

\bibitem[Banomyong et~al., 2019]{banomyong2019systematic}
Banomyong, R., Varadejsatitwong, P., and Oloruntoba, R. (2019).
\newblock A systematic review of humanitarian operations, humanitarian logistics and humanitarian supply chain performance literature 2005 to 2016.
\newblock {\em Annals of Operations Research}, 283:71--86.

\bibitem[Bayram, 2016]{BAYRAM201663}
Bayram, V. (2016).
\newblock Optimization models for large scale network evacuation planning and management: A literature review.
\newblock {\em Surveys in Operations Research and Management Science}, 21(2):63--84.

\bibitem[Ben-Tal et~al., 2011]{BENTAL20111177}
Ben-Tal, A., Chung, B.~D., Mandala, S.~R., and Yao, T. (2011).
\newblock Robust optimization for emergency logistics planning: Risk mitigation in humanitarian relief supply chains.
\newblock {\em Transportation Research Part B: Methodological}, 45(8):1177--1189.
\newblock Supply chain disruption and risk management.

\bibitem[Bertsimas et~al., 2021]{bertsimas2021predictions}
Bertsimas, D., Boussioux, L., Cory-Wright, R., Delarue, A., Digalakis, V., Jacquillat, A., Kitane, D.~L., Lukin, G., Li, M., Mingardi, L., et~al. (2021).
\newblock From predictions to prescriptions: A data-driven response to covid-19.
\newblock {\em Health care management science}, 24:253--272.

\bibitem[Bi et~al., 2023]{bi2023accurate}
Bi, K., Xie, L., Zhang, H., Chen, X., Gu, X., and Tian, Q. (2023).
\newblock Accurate medium-range global weather forecasting with 3d neural networks.
\newblock {\em Nature}, pages 1--6.

\bibitem[Boccardo et~al., 2015]{boccardo2015uav}
Boccardo, P., Chiabrando, F., Dutto, F., Giulio~Tonolo, F., and Lingua, A. (2015).
\newblock Uav deployment exercise for mapping purposes: Evaluation of emergency response applications.
\newblock {\em Sensors}, 15(7):15717--15737.

\bibitem[Boonmee et~al., 2017]{BOONMEE2017485}
Boonmee, C., Arimura, M., and Asada, T. (2017).
\newblock Facility location optimization model for emergency humanitarian logistics.
\newblock {\em International Journal of Disaster Risk Reduction}, 24:485--498.

\bibitem[Bouchard et~al., 2022]{build2}
Bouchard, I., Rancourt, M.-{\`E}., Aloise, D., and Kalaitzis, F. (2022).
\newblock On transfer learning for building damage assessment from satellite imagery in emergency contexts.
\newblock {\em Remote Sensing}, 14(11):2532.

\bibitem[Boussioux et~al., 2022]{boussioux2022hurricane}
Boussioux, L., Zeng, C., Gu{\'e}nais, T., and Bertsimas, D. (2022).
\newblock Hurricane forecasting: A novel multimodal machine learning framework.
\newblock {\em Weather and Forecasting}, 37(6):817--831.

\bibitem[Britannica, 2023]{britannica}
Britannica, E. (2023).
\newblock Morocco earthquake of 2023.

\bibitem[Burke et~al., 2019]{burke2019thermal}
Burke, C., Wich, S., Kusin, K., McAree, O., Harrison, M.~E., Ripoll, B., and Ermiasi, Y. (2019).
\newblock Thermal-drones as a safe and reliable method for detecting subterranean peat fires.
\newblock {\em Drones}, 3(1).

\bibitem[{Center for Disaster Philanthropy}, 2023]{disPhil}
{Center for Disaster Philanthropy} (2023).
\newblock 2023 morocco earthquake.

\bibitem[Chamola et~al., 2020]{chamola2020disaster}
Chamola, V., Hassija, V., Gupta, S., Goyal, A., Guizani, M., and Sikdar, B. (2020).
\newblock Disaster and pandemic management using machine learning: a survey.
\newblock {\em IEEE Internet of Things Journal}, 8(21):16047--16071.

\bibitem[Chang and Chang, 2006]{chang2006adaptive}
Chang, F.-J. and Chang, Y.-T. (2006).
\newblock Adaptive neuro-fuzzy inference system for prediction of water level in reservoir.
\newblock {\em Advances in water resources}, 29(1):1--10.

\bibitem[Chelidze et~al., 2022]{Chelidze2022}
Chelidze, T., Kiria, T., Melikadze, G., Jimsheladze, T., and Kobzev, G. (2022).
\newblock Earthquake forecast as a machine learning problem for imbalanced datasets: Example of georgia, caucasus.
\newblock {\em Frontiers in Earth Science}, 10.

\bibitem[Chen et~al., 2020]{atmos11070676}
Chen, R., Zhang, W., and Wang, X. (2020).
\newblock Machine learning in tropical cyclone forecast modeling: A review.
\newblock {\em Atmosphere}, 11(7).

\bibitem[Cheng et~al., 2021]{cheng2021deep}
Cheng, C.-S., Behzadan, A.~H., and Noshadravan, A. (2021).
\newblock Deep learning for post-hurricane aerial damage assessment of buildings.
\newblock {\em Computer-Aided Civil and Infrastructure Engineering}, 36(6):695--710.

\bibitem[Commission, 2023]{european_commission_2023}
Commission, E. (2023).
\newblock Drones and planes: unprecedented imagery resolution supports disaster assessment.
\newblock EU Science Hub.

\bibitem[Correia et~al., 2021]{9476381}
Correia, A., Água, P.~B., and Luzes, T. (2021).
\newblock Virtual reality for rescue operations training.
\newblock In {\em 2021 16th Iberian Conference on Information Systems and Technologies (CISTI)}, pages 1--6.

\bibitem[Daud et~al., 2022]{daud2022applications}
Daud, S. M. S.~M., Yusof, M. Y. P.~M., Heo, C.~C., Khoo, L.~S., Singh, M. K.~C., Mahmood, M.~S., and Nawawi, H. (2022).
\newblock Applications of drone in disaster management: A scoping review.
\newblock {\em Science \& Justice}, 62(1):30--42.

\bibitem[Deepglobe.org, 2023]{deepglobeorg2023deepglobe}
Deepglobe.org (2023).
\newblock Deepglobe - cvpr18 - home.
\newblock Accessed 2023-09-28.

\bibitem[DHS Science \& Technology Press~Office, 2022]{DHSNASA}
DHS Science \& Technology Press~Office, J.~V. (2022).
\newblock Dhs and nasa technology helps save four in nepal earthquake disaster.
\newblock \url{https://www.jpl.nasa.gov/news/dhs-and-nasa-technology-helps-save-four-in-nepal-earthquake-disaster}.

\bibitem[D{\"o}yen et~al., 2012]{doyen2012two}
D{\"o}yen, A., Aras, N., and Barbaroso{\u{g}}lu, G. (2012).
\newblock A two-echelon stochastic facility location model for humanitarian relief logistics.
\newblock {\em Optimization Letters}, 6:1123--1145.

\bibitem[Duarte et~al., 2018]{duarte2018satellite}
Duarte, D., Nex, F., Kerle, N., and Vosselman, G. (2018).
\newblock Satellite image classification of building damages using airborne and satellite image samples in a deep learning approach.
\newblock {\em ISPRS Annals of the Photogrammetry and Remote Sensing}, 4:89--96.

\bibitem[Espeholt et~al., 2022]{espeholt2022deep}
Espeholt, L., Agrawal, S., S{\o}nderby, C., Kumar, M., Heek, J., Bromberg, C., Gazen, C., Carver, R., Andrychowicz, M., Hickey, J., et~al. (2022).
\newblock Deep learning for twelve hour precipitation forecasts.
\newblock {\em Nature communications}, 13(1):1--10.

\bibitem[Farahani et~al., 2020]{FARAHANI2020787}
Farahani, R.~Z., Lotfi, M., Baghaian, A., Ruiz, R., and Rezapour, S. (2020).
\newblock Mass casualty management in disaster scene: A systematic review of {OR\&MS} research in humanitarian operations.
\newblock {\em European Journal of Operational Research}, 287(3):787--819.

\bibitem[Gaanoun et~al., 2023]{darijabert}
Gaanoun, K., Naira, A.~M., Allak, A., and Benelallam, I. (2023).
\newblock Darijabert: a step forward in nlp for the written moroccan dialect.

\bibitem[Gao et~al., 2023]{gao2023hurricast}
Gao, S., Gao, M., Li, Y., and Dong, W. (2023).
\newblock Hurricast: An automatic framework using machine learning and statistical modeling for hurricane forecasting.
\newblock {\em arXiv preprint arXiv:2309.07174}.

\bibitem[Gevaert et~al., 2021]{gevaert2021fairness}
Gevaert, C., Carman, M., Rosman, B., Georgiadou, Y., and Soden, R. (2021).
\newblock Fairness and accountability of ai in disaster risk management: Opportunities and challenges.
\newblock {\em Patterns}, 2(11).

\bibitem[GFDRR, 2018]{gfdrr2018machine}
GFDRR (2018).
\newblock {\em Machine Learning for Disaster Management}.
\newblock GFDRR, Washington DC.

\bibitem[Ghasemi et~al., 2020]{GHASEMI2020100745}
Ghasemi, P., Khalili-Damghani, K., Hafezalkotob, A., and Raissi, S. (2020).
\newblock Stochastic optimization model for distribution and evacuation planning (a case study of tehran earthquake).
\newblock {\em Socio-Economic Planning Sciences}, 71:100745.

\bibitem[Gholami-Zanjani et~al., 2019]{gholami2019dynamic}
Gholami-Zanjani, S.~M., Jafari-Marandi, R., Pishvaee, M.~S., and Klibi, W. (2019).
\newblock Dynamic vehicle routing problem with cooperative strategy in disaster relief.
\newblock {\em International Journal of Shipping and Transport Logistics}, 11(6):455--475.

\bibitem[GitHub, 2023]{github2023ultralyticsultralytics}
GitHub (2023).
\newblock ultralytics/ultralytics: New - yolov8 in pytorch > onnx > openvino > coreml > tflite.
\newblock Accessed 2023-09-28.

\bibitem[Gupta et~al., 2019a]{xbd}
Gupta, R., Hosfelt, R., Sajeev, S., Patel, N., Goodman, B., Doshi, J., Heim, E., Choset, H., and Gaston, M. (2019a).
\newblock xbd: A dataset for assessing building damage from satellite imagery.
\newblock {\em arXiv preprint arXiv:1911.09296}.

\bibitem[Gupta et~al., 2019b]{gupta2019big}
Gupta, S., Altay, N., and Luo, Z. (2019b).
\newblock Big data in humanitarian supply chain management: A review and further research directions.
\newblock {\em Annals of Operations Research}, 283:1153--1173.

\bibitem[Huggingface.co, 2023]{huggingfaceco2023humandetection}
Huggingface.co (2023).
\newblock Humandetection - a hugging face space by pkaushik.
\newblock Accessed 2023-09-30.

\bibitem[Huot et~al., 2022]{huot2022next}
Huot, F., Hu, R.~L., Goyal, N., Sankar, T., Ihme, M., and Chen, Y.-F. (2022).
\newblock Next day wildfire spread: A machine learning dataset to predict wildfire spreading from remote-sensing data.
\newblock {\em IEEE Transactions on Geoscience and Remote Sensing}, 60:1--13.

\bibitem[{International Medical Corps}, 2023]{imc}
{International Medical Corps} (2023).
\newblock Morocco earthquake: Situation report \#4.
\newblock \url{https://reliefweb.int/report/morocco/morocco-earthquake-ibc-situation-report-22-september-2023}.

\bibitem[Ivić, 2019]{ivic2019artificial}
Ivić, M. (2019).
\newblock Artificial intelligence and geospatial analysis in disaster management.
\newblock {\em The International Archives of the Photogrammetry, Remote Sensing and Spatial Information Sciences}, XLII(3).

\bibitem[Jayakumar et~al., 2020]{jayakumar2020wildfire}
Jayakumar, A., Shaji, A., and Nitha, L. (2020).
\newblock Wildfire forecast within the districts of kerala using fuzzy and anfis.
\newblock In {\em 2020 Fourth International Conference on Computing Methodologies and Communication (ICCMC)}, pages 666--669. IEEE.

\bibitem[Jia et~al., 2007]{JIA2007257}
Jia, H., Ordóñez, F., and Dessouky, M.~M. (2007).
\newblock Solution approaches for facility location of medical supplies for large-scale emergencies.
\newblock {\em Computers \& Industrial Engineering}, 52(2):257--276.

\bibitem[Jimeno-S{\'a}ez et~al., 2017]{jimeno2017estimation}
Jimeno-S{\'a}ez, P., Senent-Aparicio, J., P{\'e}rez-S{\'a}nchez, J., Pulido-Velazquez, D., and Cecilia, J.~M. (2017).
\newblock Estimation of instantaneous peak flow using machine-learning models and empirical formula in peninsular spain.
\newblock {\em Water}, 9(5):347.

\bibitem[Kia et~al., 2012]{kia2012artificial}
Kia, M.~B., Pirasteh, S., Pradhan, B., Rodzi, A., and Moradi, A. (2012).
\newblock An artificial neural network model for flood simulation using gis: Johor river basin, malaysia.
\newblock {\em Environmental Earth Sciences}, 67:251--264.

\bibitem[Klibi et~al., 2018]{klibi2018prepositioning}
Klibi, W., Ichoua, S., and Martel, A. (2018).
\newblock Prepositioning emergency supplies to support disaster relief: a case study using stochastic programming.
\newblock {\em INFOR: Information Systems and Operational Research}, 56(1):50--81.

\bibitem[Kuglistsch et~al., 2022]{kuglistsch2022facilitating}
Kuglistsch, M.~M., Pelivan, I., Ceola, S., Menon, M., and Xoplaki, E. (2022).
\newblock Facilitating adoption of ai in natural disaster management through collaboration.
\newblock {\em Nature Communications}, 13.

\bibitem[Kuglitsch et~al., 2022a]{MR4}
Kuglitsch, M., Albayrak, A., Aquino, R., Craddock, A., Edward-Gill, J., Kanwar, R., Koul, A., Ma, J., Marti, A., Menon, M., et~al. (2022a).
\newblock Artificial intelligence for disaster risk reduction: Opportunities, challenges, and prospects.
\newblock {\em Bulletin n{\textordmasculine}}, 71(1).

\bibitem[Kuglitsch et~al., 2022b]{MR3}
Kuglitsch, M.~M., Pelivan, I., Ceola, S., Menon, M., and Xoplaki, E. (2022b).
\newblock Facilitating adoption of ai in natural disaster management through collaboration.
\newblock {\em Nature communications}, 13(1):1579.

\bibitem[Kundu et~al., 2022]{kundu2022emergency}
Kundu, T., Sheu, J.-B., and Kuo, H.-T. (2022).
\newblock Emergency logistics management—review and propositions for future research.
\newblock {\em Transportation research part E: logistics and transportation review}, 164:102789.

\bibitem[Kunz et~al., 2017]{kunz2017relevance}
Kunz, N., Van~Wassenhove, L.~N., Besiou, M., Hambye, C., and Kovacs, G. (2017).
\newblock Relevance of humanitarian logistics research: best practices and way forward.
\newblock {\em International Journal of Operations \& Production Management}, 37(11):1585--1599.

\bibitem[Lam et~al., 2022]{lam2022graphcast}
Lam, R., Sanchez-Gonzalez, A., Willson, M., Wirnsberger, P., Fortunato, M., Pritzel, A., Ravuri, S., Ewalds, T., Alet, F., Eaton-Rosen, Z., et~al. (2022).
\newblock Graphcast: Learning skillful medium-range global weather forecasting.
\newblock {\em arXiv preprint arXiv:2212.12794}.

\bibitem[Lane and York, 2022]{lane-2023-naturaldisasters}
Lane, L. and York, H. (2022).
\newblock How natural disasters exacerbate human trafficking.

\bibitem[Li et~al., 2022]{li2022yolov6}
Li, C., Li, L., Jiang, H., Weng, K., Geng, Y., Li, L., Ke, Z., Li, Q., Cheng, M., Nie, W., et~al. (2022).
\newblock Yolov6: A single-stage object detection framework for industrial applications.
\newblock {\em arXiv preprint arXiv:2209.02976}.

\bibitem[Maxar.com, 2023]{maxarcom2023morocco}
Maxar.com (2023).
\newblock Morocco earthquake september 2023 | maxar.
\newblock Accessed 2023-09-28.

\bibitem[Michelini et~al., 2021]{michelini2021instance}
Michelini, A., Cianetti, S., Gaviano, S., Giunchi, C., Jozinovi{\'c}, D., and Lauciani, V. (2021).
\newblock Instance--the italian seismic dataset for machine learning.
\newblock {\em Earth System Science Data}, 13(12):5509--5544.

\bibitem[Mignan and Broccardo, 2020]{Mignan2020}
Mignan, A. and Broccardo, M. (2020).
\newblock Neural network applications in earthquake prediction (1994–2019): Meta‐analytic and statistical insights on their limitations.
\newblock {\em Seismological Research Letters}, 91(4):2330--2342.

\bibitem[Mosavi et~al., 2018]{mosavi2018flood}
Mosavi, A., Ozturk, P., and Chau, K.-w. (2018).
\newblock Flood prediction using machine learning models: Literature review.
\newblock {\em Water}, 10(11):1536.

\bibitem[Munawar et~al., 2021a]{munawar2021integrated}
Munawar, H.~S., Hammad, A.~W., Waller, S.~T., Thaheem, M.~J., and Shrestha, A. (2021a).
\newblock An integrated approach for post-disaster flood management via the use of cutting-edge technologies and uavs: A review.
\newblock {\em Sustainability}, 13(14):7925.

\bibitem[Munawar et~al., 2021b]{munawar2021application}
Munawar, H.~S., Ullah, F., Qayyum, S., and Heravi, A. (2021b).
\newblock Application of deep learning on uav-based aerial images for flood detection.
\newblock {\em Smart Cities}, 4(3):1220--1242.

\bibitem[Narayan, 2021]{Narayan2018}
Narayan, Y. (2021).
\newblock Deepquake: Artificial intelligence for earthquake forecasting using fine-grained climate data.
\newblock In {\em NeurIPS 2021 Workshop on Tackling Climate Change with Machine Learning}.

\bibitem[of~Homeland~Security, 2022]{DHS}
of~Homeland~Security, U.~D. (2022).
\newblock Detecting heartbeats in rubble: Dhs and nasa team up to save victims of disasters.
\newblock \url{https://www.dhs.gov/detecting-heartbeats-rubble-dhs-and-nasa-team-save-victims-disasters}.

\bibitem[Pathak et~al., 2022]{pathak2022fourcastnet}
Pathak, J., Subramanian, S., Harrington, P., Raja, S., Chattopadhyay, A., Mardani, M., Kurth, T., Hall, D., Li, Z., Azizzadenesheli, K., Hassanzadeh, P., Kashinath, K., and Anandkumar, A. (2022).
\newblock Four{C}ast{N}et: A {G}lobal {D}ata-driven {H}igh-resolution {W}eather {M}odel using {A}daptive {F}ourier {N}eural {O}perators.

\bibitem[P{\'e}rez-Porras et~al., 2021]{perez2021machine}
P{\'e}rez-Porras, F.-J., Trivi{\~n}o-Tarradas, P., Cima-Rodr{\'\i}guez, C., Mero{\~n}o-de Larriva, J.-E., Garc{\'\i}a-Ferrer, A., and Mesas-Carrascosa, F.-J. (2021).
\newblock Machine learning methods and synthetic data generation to predict large wildfires.
\newblock {\em Sensors}, 21(11):3694.

\bibitem[Radke et~al., 2019]{radke2019firecast}
Radke, D., Hessler, A., and Ellsworth, D. (2019).
\newblock Firecast: Leveraging deep learning to predict wildfire spread.
\newblock In {\em IJCAI}, pages 4575--4581.

\bibitem[Rana et~al., 2022]{Rana2015}
Rana, A., Vaidya, P., and Hu, Y.-C. (2022).
\newblock A comparative analysis of ann and anfis approaches for earthquake forecasting.
\newblock In {\em 2022 Second International Conference on Computer Science, Engineering and Applications (ICCSEA)}, pages 1--6. IEEE.

\bibitem[Reyes et~al., 2013]{Reyes2013}
Reyes, J., Morales-Esteban, A., and Martínez-Álvarez, F. (2013).
\newblock Neural networks to predict earthquakes in chile.
\newblock {\em Applied Soft Computing Journal}, 13:1314--1328.

\bibitem[Rezaeian~Zadeh et~al., 2010]{rezaeian2010daily}
Rezaeian~Zadeh, M., Amin, S., Khalili, D., and Singh, V.~P. (2010).
\newblock Daily outflow prediction by multi layer perceptron with logistic sigmoid and tangent sigmoid activation functions.
\newblock {\em Water resources management}, 24:2673--2688.

\bibitem[Saad et~al., 2023]{Saad2018}
Saad, O.~M., Chen, Y., Savvaidis, A., Fomel, S., Jiang, X., Huang, D., Obou{\'e}, Y. A. S.~I., Yong, S., Wang, X., Zhang, X., et~al. (2023).
\newblock Earthquake forecasting using big data and artificial intelligence: A 30-week real-time case study in china.
\newblock {\em Bull. Seismol. Soc. Am}, 20:1--18.

\bibitem[Shiri et~al., 2020]{9169921}
Shiri, H., Park, J., and Bennis, M. (2020).
\newblock Communication-efficient massive uav online path control: Federated learning meets mean-field game theory.
\newblock {\em IEEE Transactions on Communications}, 68(11):6840--6857.

\bibitem[Snezhana, 2023]{Dineva2023}
Snezhana, D. (2023).
\newblock Applying artificial intelligence (ai) for mitigation climate change consequences of the natural disasters.
\newblock {\em Dineva, S.(2023). Applying Artificial Intelligence (AI) for Mitigation Climate Change Consequences of the Natural Disasters. Research Journal of Ecology and Environmental Sciences}, 3(1):1--8.

\bibitem[S{\o}nderby et~al., 2020]{sonderby2020metnet}
S{\o}nderby, C.~K., Espeholt, L., Heek, J., Dehghani, M., Oliver, A., Salimans, T., Agrawal, S., Hickey, J., and Kalchbrenner, N. (2020).
\newblock Metnet: A neural weather model for precipitation forecasting.
\newblock {\em arXiv preprint arXiv:2003.12140}.

\bibitem[Stienen et~al., 2021]{STIENEN2021102494}
Stienen, V., Wagenaar, J., {den Hertog}, D., and Fleuren, H. (2021).
\newblock Optimal depot locations for humanitarian logistics service providers using robust optimization.
\newblock {\em Omega}, 104:102494.

\bibitem[Su et~al., 2020]{build1}
Su, J., Bai, Y., Wang, X., Lu, D., Zhao, B., Yang, H., Mas, E., and Koshimura, S. (2020).
\newblock Technical solution discussion for key challenges of operational convolutional neural network-based building-damage assessment from satellite imagery: Perspective from benchmark xbd dataset.
\newblock {\em Remote Sensing}, 12(22):3808.

\bibitem[Sun et~al., 2020]{sun2020a}
Sun, W., Bocchini, P., and Davison, B.~D. (2020).
\newblock {Applications of artificial intelligence for disaster management}.
\newblock {\em Natural Hazards}, 103(3):2631--2689.

\bibitem[Syifa et~al., 2019]{syifa2019artificial}
Syifa, M., Kadavi, P.~R., and Lee, C.-W. (2019).
\newblock An artificial intelligence application for post-earthquake damage mapping in palu, central sulawesi, indonesia.
\newblock {\em Sensors}, 19(3).

\bibitem[Tan et~al., 2021]{MR2}
Tan, L., Guo, J., Mohanarajah, S., and Zhou, K. (2021).
\newblock Can we detect trends in natural disaster management with artificial intelligence? a review of modeling practices.
\newblock {\em Natural Hazards}, 107:2389--2417.

\bibitem[Unosat.org, 2023]{unosatorg2023unosat}
Unosat.org (2023).
\newblock Unosat.
\newblock Accessed 2023-09-28.

\bibitem[Velev and Zlateva, 2023]{Velev2023}
Velev, D. and Zlateva, P. (2023).
\newblock {Challenges of Artificial Intelligence Application for Disaster Risk Management}.
\newblock {\em International Archives of the Photogrammetry, Remote Sensing and Spatial Information Sciences - ISPRS Archives}, 48(M-1-2023):387--394.

\bibitem[Verdecchia et~al., 2023]{verdecchia2023systematic}
Verdecchia, R., Sallou, J., and Cruz, L. (2023).
\newblock A systematic review of green ai.
\newblock {\em Wiley Interdisciplinary Reviews: Data Mining and Knowledge Discovery}, page e1507.

\bibitem[Wu et~al., 2021]{wu2021building}
Wu, C., Zhang, F., Xia, J., Xu, Y., Li, G., Xie, J., Du, Z., and Liu, R. (2021).
\newblock Building damage detection using u-net with attention mechanism from pre-and post-disaster remote sensing datasets.
\newblock {\em Remote Sensing}, 13(5):905.

\bibitem[Xiang et~al., 2020]{xiang2020rainfall}
Xiang, Z., Yan, J., and Demir, I. (2020).
\newblock A rainfall-runoff model with lstm-based sequence-to-sequence learning.
\newblock {\em Water resources research}, 56(1):e2019WR025326.

\bibitem[Yigitcanlar, 2021]{Tan2020}
Yigitcanlar, T. (2021).
\newblock Greening the artificial intelligence for a sustainable planet: An editorial commentary.

\bibitem[Yu et~al., 2017]{yu2017comparison}
Yu, P.-S., Yang, T.-C., Chen, S.-Y., Kuo, C.-M., and Tseng, H.-W. (2017).
\newblock Comparison of random forests and support vector machine for real-time radar-derived rainfall forecasting.
\newblock {\em Journal of hydrology}, 552:92--104.

\bibitem[Zhao et~al., 2022]{zhao2022road}
Zhao, K., Liu, J., Wang, Q., Wu, X., and Tu, J. (2022).
\newblock Road damage detection from post-disaster high-resolution remote sensing images based on tld framework.
\newblock {\em IEEE Access}, 10:43552--43561.

\bibitem[Zhu et~al., 2023]{Zhu2023}
Zhu, C., Cotton, F., Kawase, H., and Nakano, K. (2023).
\newblock How well can we predict earthquake site response so far? machine learning vs physics-based modeling.
\newblock {\em Earthquake Spectra}, 39:478--504.

\bibitem[Zhu et~al., 2021]{zhu2021msnet}
Zhu, X., Liang, J., and Hauptmann, A. (2021).
\newblock Msnet: A multilevel instance segmentation network for natural disaster damage assessment in aerial videos.
\newblock In {\em Proceedings of the IEEE/CVF winter conference on applications of computer vision}, pages 2023--2032.

\end{thebibliography}

\appendix

\newpage
\section*{Morocco Solidarity Hackathon }
\begin{multicols}{2}
\noindent
\textbf{Organizing team}
\begin{itemize}
  \item Léna Néhale Ezzine \smallts{1, 2} \textsuperscript{$\ast$}
  \item Ayoub Atanane \smallts{1, 3}
  \item Ghait Boukachab \smallts{1}
  \item Oussama Boussif \smallts{1, 2}
  \item Mohammed Mahfoud \smallts{1, 4}
  \item Yassine Yaakoubi \smallts{5}
  \item Loubna Benabbou \smallts{2, 3}
\end{itemize}
\textbf{Speakers}
\begin{itemize}
  \item Yoshua Bengio \smallts{1, 2}
  \item Léonard Boussioux \smallts{6}
  \item Dick Den Hertog \smallts{7}
  \item Mehdi Bennis \smallts{8}
  \item Peetak Mitra \smallts{9}
  \item Alexandre Jacquillat \smallts{10}
  \item Omar El Housni \smallts{11}

\end{itemize}
\columnbreak
\textbf{Judges and Mentors}
\begin{itemize}
  \item Nouamane Tazi \smallts{15}
  \item Salim Chemlal \smallts{12}
  \item Wale Akinfaderin \smallts{13}
  \item Laurent Barcelo \smallts{14}
  \item Victor Schmidt \smallts{1, 2}
  \item Zhor Khadir \smallts{2}
  \item Jeremy Pinto \smallts{1}
  \item Tariq Daouda \smallts{16}
  \item Redouane Lguensat \smallts{17}
  \item Alex Hernandez-Garcia \smallts{1}
  \item Reda Snaiki \smallts{18}
  \item Syed Aamir Aarfi \smallts{13}
  \item Aboujihad Dhimine \smallts{13}
  \item Abderrahim Khalifa \smallts{13}
  \item Hamza Azzaoui \smallts{13}
  \item Dolly Akiki \smallts{13}
  \item David Min \smallts{13}
  \item Karim Bouzoubaa \smallts{31, 32}
\end{itemize}
\end{multicols}

\textbf{Teams}
\begin{multicols}{2}
\noindent
\begin{enumerate}
    \item \textbf{DeepAster}
        \begin{itemize}
          \item Ilham EL Bouloumi \smallts{19}
          \item Ayoub Loudyi \smallts{20}
          \item Aymane El Firdoussi \smallts{21}
          \item Achraf Sbai \smallts{22}
          \item Sanae Attak \smallts{16}
        \end{itemize}
    \item \textbf{SOS Team}
        \begin{itemize}
          \item Kaoutar Lakdim \smallts{24}
          \item Yassine Squalli Houssaini \smallts{24}
          \item Firdawse Guerbouzi \smallts{24}
          \item Chaimae Biyaye \smallts{24}
          \item Khadija Bayoud \smallts{24}
          \item Ikram Belmadani \smallts{24}
        \end{itemize}
    \item \textbf{Team of 5}
        \begin{itemize}
          \item Charles Bricout \smallts{1}
          \item Alex Maggioni \smallts{1}
          \item Reyad OUAHI \smallts{25}
          \item B.V. Alaka \smallts{26}
          \item Kiruthika Subramani \smallts{27}
    \end{itemize}
\end{enumerate}
\columnbreak
\begin{enumerate}
\setcounter{enumi}{3}
      \item \textbf{Groomin in Darija}
        \begin{itemize}
          \item Khaoula Chehbouni \smallts{1, 5}
          \item Afaf Taik \smallts{1}
          \item Kanishk Jain \smallts{28}
        \end{itemize}
        \item \textbf{Datacuration}
            \begin{itemize}
              \item Hamza Ghernati \smallts{29}
              \item Lamia Salhi \smallts{30}
              \item Laila Salhi \smallts{31}
              \item Jules Lambert
        \end{itemize}
        \item \textbf{ML4Quake}
            \begin{itemize}
              \item Yuyan Chen \smallts{1, 5}
              \item Nikhil Reddy Pottanigari \smallts{1, 2}
              \item Santhoshi Ravichandran \smallts{1, 2}
              \item Ashwini Rajaram \smallts{1, 2}
            \end{itemize}
\end{enumerate}
\end{multicols}

\footnote[1]{Corresponding authors: lena-nehale.ezzine@mila.quebec}

\newpage
\subsection*{Affiliations}
\begin{multicols}{2}
\begin{enumerate}
  \item Mila - Québec AI Institute
  \item Université de Montréal
  \item Université du Québec à Rimouski
  \item Technical University of Munich
  \item McGill University
  \item University of Washington
  \item University of Amsterdam
  \item University of Oulu
  \item Excarta
  \item Massachusetts Institute of Technology
  \item Cornell University
  \item MoroccoAI
  \item Amazon Web Services
  \item Videns Analytics
  \item Hugging Face
  \item University Mohammed VI Polytechnic
\end{enumerate}
\columnbreak
\begin{enumerate}
\setcounter{enumi}{16}
  \item Institut Pierre-Simon Laplace
  \item École de Technologie Supérieure
  \item DataChainEd
  \item INP
  \item Télécom Paris 
  \item Inria
  \item UEMF
  \item ENSIAS
  \item ML Collective
  \item M Kumarasamy College of Engineering 
  \item IIT Hyderabad 
  \item Montréal International 
  \item Ohmic technologies 
  \item DentalMonitoring
  \item Mohammadia School of Engineers
  \item Mohammed Vth University in Rabat

\end{enumerate}
\end{multicols}

\newpage
\begin{center}
    \begin{Large}
    \textbf{Supplementary Material}
    \end{Large}
\end{center}


\section{Quality Evaluation Criteria of Proposals}\label{judges_criteria}
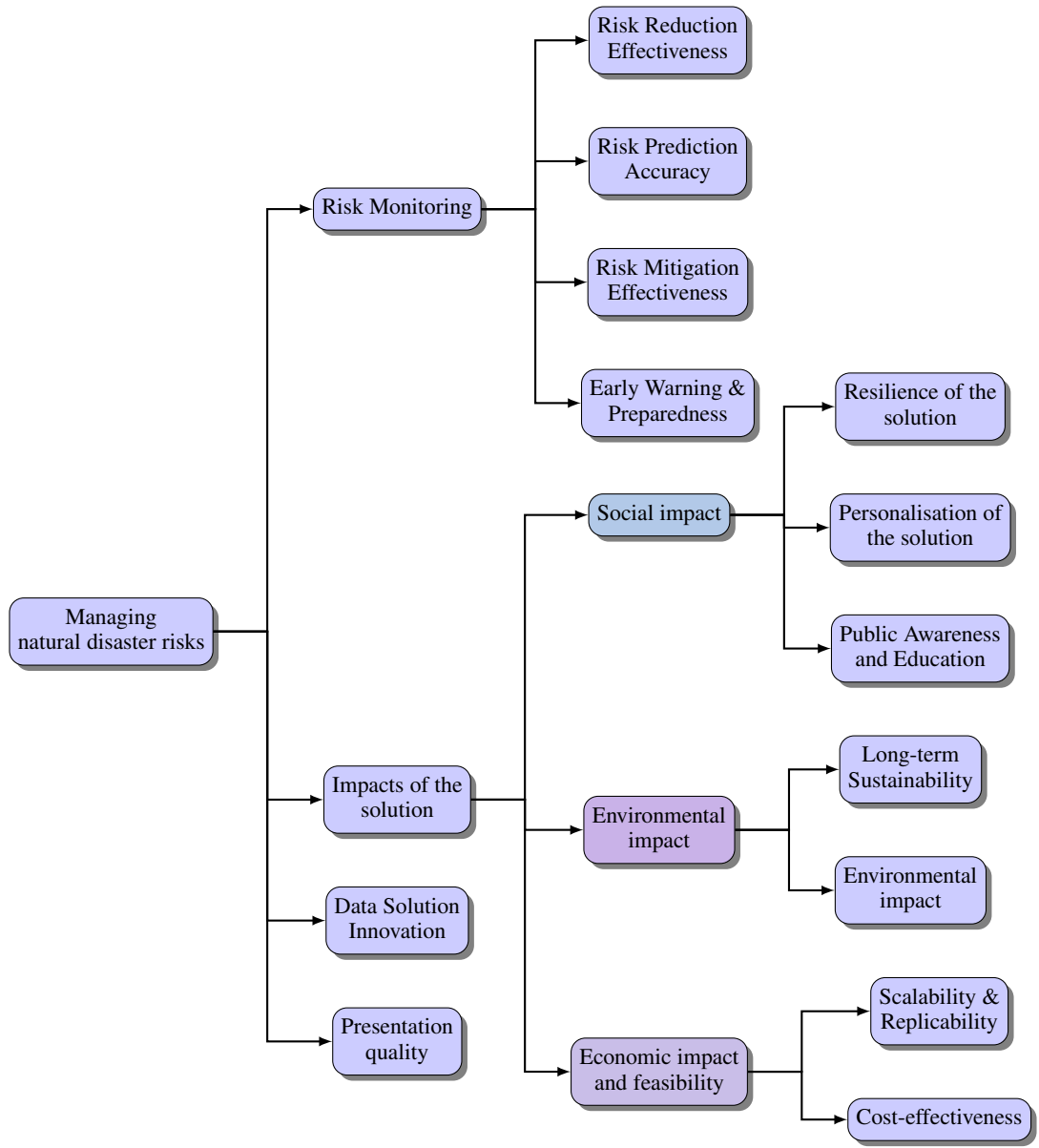
\begin{figure}[ht]
    \centering
    \begin{forest}
      for tree={
        grow'=0,
        draw,
        rounded corners=2mm,
        fill=blue!20,
        font=\footnotesize,
        align=center,
        drop shadow,
        parent anchor=east,
        child anchor=west,
        edge={thick, -latex},
        l sep+=1cm,
        s sep+=0.5cm,
        edge path={
          \noexpand\path[\forestoption{edge}]
          (!u.parent anchor) -- +(0.75cm,0) |- (.child anchor)\forestoption{edge label};
        }, 
      }
      [Managing\\ natural disaster risks
        [Risk Monitoring
          [Risk Reduction\\ Effectiveness]
          [Risk Prediction\\ Accuracy]
          [Risk Mitigation\\ Effectiveness]
          [Early Warning \&\\ Preparedness]
        ]
        [Impacts of the\\ solution
          [Social impact, fill=green!30!blue!30
            [Resilience of the\\ solution]
            [Personalisation of\\ the solution]
            [Public Awareness\\ and Education]
          ]
          [Environmental\\ impact, fill=red!30!blue!30
            [Long-term\\ Sustainability]
            [Environmental\\ impact]
          ]
          [Economic impact\\ and feasibility, fill=orange!30!blue!30
            [Scalability \&\\ Replicability]
            [Cost-effectiveness]
          ]
        ]
        [Data Solution\\ Innovation]
        [Presentation\\ quality]
      ]
    \end{forest}
    \caption{Criteria deployed during the Moroccan Solidarity Hackathon to assess the teams' proposals.}
    \label{fig:tree_diagram}
\end{figure}





\end{document}